\documentclass[10pt, a4paper]{article}
\usepackage{multirow}
\usepackage{enumitem}
\usepackage{float}

\usepackage[final]{lrec2026} 

\title{An Exploration-Analysis-Disambiguation Reasoning Framework for Word Sense Disambiguation with Low-Parameter LLMs}

\name{Deshan Sumanathilaka, Nicholas Micallef, Julian Hough} 

\address{Department of Computer Science, Swansea University \\
         Fabian Way, Crymlyn Burrows, Swansea, Wales, UK \\         
         \{t.g.d.sumanathilaka, nicholas.micallef, julian.hough\}@swansea.ac.uk\\}

\abstract{
Word Sense Disambiguation (WSD) remains a key challenge in Natural Language Processing (NLP), especially when dealing with rare or domain-specific senses that are often misinterpreted. While modern high-parameter Large Language Models (LLMs) such as GPT-4-Turbo have shown state-of-the-art WSD performance, their computational and energy demands limit scalability. This study investigates whether low-parameter LLMs (\texttt{<}4B parameters) can achieve comparable results through fine-tuning strategies that emphasize reasoning-driven sense identification. Using the FEWS dataset augmented with semi-automated, rationale-rich annotations, we fine-tune eight small-scale open-source LLMs (e.g.  Gemma and Qwen). Our results reveal that Chain-of-Thought (CoT)-based reasoning combined with neighbour-word analysis achieves performance comparable to GPT-4-Turbo in zero-shot settings. Importantly, Gemma-3-4B and Qwen-3-4B models consistently outperform all medium-parameter baselines and state-of-the-art models on FEWS, with robust generalization to unseen senses. Furthermore, evaluation on the unseen ``Fool Me If You Can'' dataset confirms strong cross-domain adaptability without task-specific fine-tuning. This work demonstrates that with carefully crafted reasoning-centric fine-tuning, low-parameter LLMs can deliver accurate WSD while substantially reducing computational and energy demands.
 \\ \newline 
 \Keywords{Word Sense Disambiguation, Low-parameter LLMs, Reasoning-driven Fine-tuning}}

\begin{document}

\maketitleabstract

\section{Introduction}

With recent advances in large language models (LLMs), Natural Language Processing (NLP) tasks such as machine translation, information retrieval, question answering, sentiment analysis, and summarization have achieved near-human performance in modern applications~\cite{chang2024survey,sumanathilaka-etal-2025-prompt}. However, to perform such linguistic tasks effectively, word sense disambiguation (WSD) plays a major role, as ambiguity can lead to incorrect information and the generation of misinformation~\cite{singh2024context}. For example, during the COVID-19 pandemic, the term ``\textit{positive}'' could indicate a confirmed infection or express optimism in a general context which are two drastically different meanings that require contextual understanding to avoid incorrect inferences. Understanding the meaning of a word in a sentence or phrase and disambiguating the correct sense is crucial for achieving accurate results, particularly when tackling the lexical ambiguity of an ambiguous word. 

Working with words that have multiple meanings (polysemy) is a major challenge for current NLP systems, as highly polysemous words are difficult to disambiguate due to the nature of usage. For example, according to BabelNet\footnote{\url{https://babelnet.org/}}, the word `\textit{bank}' has 49 meanings, including 40 as a nouns and 9 as a verb. However, in common usage, the senses ``financial institution'' (money bank) and ``sloping land beside a body of water'' (river bank) dominate. In contrast, senses including ``a flight maneuver'' (highly domain-specific to aviation) or ``a slope in the turn of a road or track'' (used mainly in engineering, road design) are rare in everyday usage. In domain-specific situations, models often struggle to disambiguate less common senses effectively~\cite{awotunde2025word}.  

Recent studies on WSD using large language models (LLMs) have revealed promising findings, including significant improvements in disambiguating rare senses and leveraging broader contextual information, paving the way for enhanced language understanding and more accurate disambiguation. For instance,~\citet{sumanathilaka_can_2024} evaluated multiple large-parameter LLMs for disambiguation using few-shot and knowledge-base-combined methods, highlighting the strengths of LLMs for WSD. However, other studies also reveal important limitations.~\citet{basile2025exploring} demonstrated that while LLMs perform well in zero-shot scenarios, they cannot surpass current state-of-the-art methods without fine-tuning, suggesting that existing approaches do not fully exploit LLM capabilities or generalize across diverse WSD settings. These gaps highlight the need for research that addresses scalability, efficiency, and more effective use of LLMs for robust and generalizable disambiguation. In particular, incorporating proper inner reasoning to guide decision making is crucial, motivating the present study on reasoning-driven WSD with low-parameter LLMs. Inspired by these observations, we design and evaluate a smaller, energy-efficient LLMs for WSD with three primary design strategies: 
\begin{enumerate}[leftmargin=*,nosep]
    \item Given a word in a sentence, the LLM must generate the correct definition.  
    \item Given a word in a sentence and possible senses, the model reasons using the neighbouring context closely related to the disambiguation process to identify the correct sense.  
    \item Given a word in a sentence and possible senses, the model reasons why a particular sense is correct and why other senses are not, to determine the correct meaning.  
\end{enumerate}

Following these design principles, we fine-tune energy-efficient, low-parameter open-source models with fewer than 4B parameters~\cite{bai2024beyond}. Our study reveals that a chain-of-thought (CoT)-based reasoning process combined with neighbouring context analysis performs comparably to large models like GPT-4-Turbo, achieving high performance on unseen data. Furthermore, our advanced reasoning approach requires only 10\% of the training data compared to the second approach, yet performed similarly, demonstrating its effectiveness. The main contributions of this work are:

\begin{itemize}[leftmargin=*,nosep]
    \item Building three reasoning datasets using a semi-automated approach and releasing them as open-source resources for future research.\footnote{Reasoning dataset: \url{https://huggingface.co/datasets/deshanksuman/Reasoning_WSD_dataset}} \footnote{Reasoning dataset for Verb: \url{https://huggingface.co/datasets/deshanksuman/Advance_Reasoning_for_Verb_WSD}}
    \item Proposing and implementing lightweight adapters following a novel EAD (Exploration, Analysis and Disambiguation) framework that can be efficiently used for WSD with small-scale models.
    \item Evaluating eight open-source small-parameter models with the proposed approaches and achieving superior performance compared to all medium-parameter models.
\end{itemize}

The rest of the paper is organized as follows: background and related work, which discusses the WSD evolution and related studies; methodology, which outlines the approach used to conduct this study; results and observations, followed by conclusions and future directions.

\section{Background and Related Work}

\subsection{Evolution of WSD}

WSD has a long history, evolving from rule-based methods to deep learning and LLM approaches. Early rule-based systems relied on hand-crafted grammatical and syntactic rules to exploit the local context of ambiguous words~\cite{bowerman1978acquisition}, but these methods were labor-intensive, domain-specific, and difficult to scale to complex languages~\cite{palmer2006evaluation}. Knowledge-based (KB) approaches emerged with resources such as WordNet, dictionaries, and thesauri incorporating algorithms such as Lesk to measure the sense overlap between gloss definitions and the surrounding context~\cite{lesk1986automatic}. Advanced KB solutions such as random walks over extended lexical graphs including Extended WordNet~\cite{agirre_random_2014} and graph-based methods that combined word embeddings with contextual information were explored~\cite{duarte_deep_2021}. Exploiting synset definitions, hypernymy relations, and contextual features improved the accuracy~\cite{kolte2009exploiting}, with frameworks like the Synset Relation-Enhanced Framework (SREF) achieving state-of-the-art KB WSD~\cite{wang_synset_2020}. Filtering unnecessary semantic information~\cite{kwon2021word} and incorporating dependency parsing to extract more precise contextual knowledge~\cite{meng2022word} were also prominent KB innovations.

With annotated corpora, supervised Machine Learning became the most common research approach~\cite{le2004high,al2016comparative,gosal2015naive}. These methods exploited features, namely target word morphology, Part of Speech (POS) tags, syntactic dependencies, and collocations. To achieve efficient WSD, supervised neural architectures incorporated richer lexical and gloss-based information to the training pipeline. Gloss-augmented neural networks jointly encoded glosses and context were examined~\cite{luo_incorporating_2018}. Other approaches to supervised WSD explored multiple-sense identification~\cite{orlando_amuse-wsd_2021}, stacked BiLSTMs with attention~\cite{laatar_evaluation_2023}, and context-dependent modeling~\cite{koppula_word_2021}.

There was a major shift in WSD research with the arrival of Transformer architectures~\cite{vaswani2017attention}, which enabled deep, attention-based contextual modeling. Transformer-based models such as BERT and GPT achieved state-of-the-art performance on WSD tasks~\cite{huang_glossbert_2020}, demonstrating strong generalization to new domains due to extensive pre-training on diverse corpora. For instance, SenseBERT augments BERT pre-training by requiring the model to predict masked words and their WordNet supersenses~\cite{levine2019sensebert}, while GlossBERT constructs context–gloss pairs and fine-tunes BERT~\cite{huang_glossbert_2020}. Moreover,~\cite{barba_consec_2021, barba_esc_2021, blevins_moving_2020} reshaped the field with supervised approaches that enhance WSD performance. 

Current research focuses on zero-shot and few-shot WSD with LLMs, applying in-context learning to leverage linguistic knowledge without extensive retraining~\cite{basile2025exploring,yae2025leveraging}. These approaches combine large-scale pre-training, dynamic context modeling, and KB augmentation, representing the most powerful and versatile WSD techniques to date and thus forming the basis of this study's methods.

\subsection{Large Language Models for WSD}

Despite their remarkable performance across numerous NLP tasks, LLMs still face several challenges in WSD. While they excel at handling common vocabulary, recent research indicates that LLMs often misinterpret rare or domain-specific ambiguous terms, especially in cross-lingual scenarios~\cite{cahyawijaya2024thank,yae2024leveraging,basile2025exploring,ortega-martin_linguistic_2023,meconi2025large}. For example,~\citet{cahyawijaya2024thank} highlighted persistent errors between languages, with a bias toward higher-resource languages. Furthermore,~\citet{yae2024leveraging} observed that model size strongly influences WSD accuracy, yet larger models demand significantly more computational resources, raising efficiency concerns. Although more parameter-efficient language models generally lack the rich contextual understanding of their larger counterparts,~\citet{basile2025exploring} demonstrates that fine-tuning them for downstream WSD tasks substantially improves accuracy over base configurations. This approach offers the potential for energy-efficient, domain-specialized disambiguation. 

Recent work has explored various strategies for leveraging LLMs in WSD.~\citet{sainz_what_2023} reframed WSD as a textual entailment task, prompting LLMs to evaluate domain label suitability for a sentence containing an ambiguous word. This zero-shot method not only outperformed random guessing but also, in certain cases, matched or exceeded supervised WSD systems~\cite{ortega-martin_linguistic_2023}. Similarly,~\citet{sumanathilaka2024assessing} investigated prompt engineering with in-context learning in GPT-3.5-Turbo and GPT-4, while subsequent work benchmarked the WSD performance of multiple LLMs, identifying DeepSeek R1 and GPT-4-mini as effective~\cite{sumanathilaka_can_2024}.

Few studies have examined the affect of functional variables in LLMs. Mainly, temperature tuning was shown to influence disambiguation accuracy~\cite{sumanathilaka2025exploring,li2025exploring}. Model architecture is crucial, with~\citet{qorib2024decoder} reporting that encoder-only models can outperform decoder-only designs for this task. Meanwhile, multilingual and translation-based WSD strategies continue to be explored~\cite{kang2023translate,david2024multilingual,ren2024few,abdel2024rematchka,laba2023contextual}, reflecting the ongoing relevance even with LLMs, as its role has shifted from a standard pipeline component to a targeted research focus.

\section{Methodology}

With the aim of achieving a reduced use of computational, memory, energy, and financial resources, we present the methodology employed in our work using low-parameter models (<4B parameters) for an efficient reasoning driven WSD task. 

\subsection{Dataset and Augmentation Process}

\begin{table}[t]
\centering
\begin{footnotesize}
\caption{Stats for fine-tuning Variants. R: Reason}
\label{tab:wsd-datasets}
\begin{tabular}{p{1.95cm}|p{0.5cm}|p{0.5cm}r|p{0.5cm}r}
\hline
\textbf{Dataset} & \textbf{Size (K)} & \multicolumn{2}{c|}{\textbf{Input}} & \multicolumn{2}{c}{\textbf{Output}} \\
\cline{3-6}
 &  & \textbf{Max} & \textbf{Avg} & \textbf{Max} & \textbf{Avg} \\
\hline
Direct sense & 101 & 511  & 44.7   & 251  & 13.9 \\
COT R & 101 & 1915 & 226.1 & 1921 & 265.9 \\
Advanced R & 10 & 1865 & 212.6  & 2477 & 672.0 \\
Verb R & 4.5 & 1915 & 222.6  & 2607 & 764.4 \\
\hline
\end{tabular}
\end{footnotesize}
\end{table}

This work employs the FEWS dataset as the basis for the experiments, which includes the sense tag list, training data, and test data~\cite{blevins_fews_2021}. FEWS was chosen because it contains less frequently used ambiguous words compared to the Semcor and Unified Evaluation Framework datasets~\cite{raganato_word_2017}. Additionally, the distribution of training and test data, designed for evaluation in both few-shot and zero-shot settings, makes it well-suited for a fair and meaningful comparison~\cite{goworek2025senwich}. The training set contains $\approx$101K samples, and each test set has 5,000 records for evaluation.


We augment the FEWS training data to support the fine-tuning process~\cite{blevins_fews_2021}. Our fine-tuning method is designed to elicit chain-of-thought (CoT) reasoning together with neighbour-word analysis to select the closest semantic candidate to determine word sense (see subsection~\ref{cot reasoning}). 
For advanced reasoning, we employ the \textit{Virtuoso-Large}, an open source model from Arcee.ai\footnote{\url{https://huggingface.co/arcee-ai/Virtuoso-Large}} to generate rationales for correct sense assignments and for rejecting competing senses. The model was selected based on its strong empirical performance reported in \citet{sumanathilaka_can_2024}, where it ranked among the best-performing models, and due to the availability of open weights, ensuring transparency and reproducibility. The model was instructed to produce a structured rationale based on three factors: contextual analysis, justification to the correct sense, and reasoning for elimination of incorrect sense. The process used a human-in-the-loop approach to ensure accurate data augmentation. We conducted additional experiments to evaluate computational techniques for verb disambiguation. To enhance the dataset, we incorporated syntactic evidence into the reasoning chain in addition to the semantic evidence used in the previous process. Furthermore, we utilized an improved prompt to generate 4.5K annotated instances for training the verb model.

The first author systematically traced and supervised the entire data generation process, ensuring a human-in-the-loop validation framework to maintain annotation quality and reliability. In addition, a representative sample of the generated data was evaluated using an LLM-as-a-judge approach \cite{li-etal-2025-generation} (See Appendix Table \ref{tab:cot-prompt-LLM-as-judge}), employing OpenAI GPT-4o and DeepSeek-V3 as independent evaluators. The evaluation yielded consistently high scores across all dimensions. DeepSeek-V3 presents a average scores of 4.906 (Contextual Analysis), 4.898 (Justification Accuracy), 4.938 (Elimination Completeness), and 4.968 (Coherence), while GPT-4o shows 4.974, 4.962, 4.964, and 4.966, respectively. These results indicate strong agreement between evaluators and confirm the high quality of the generated dataset. Table~\ref{tab:wsd-datasets} presents the statistics of the training data.

\subsection{Fine-tuning strategies}

The fine-tuning of the models was designed to address multiple objectives, allowing the evaluation of different performance criteria. Specifically, we assessed eight models to investigate whether low-parameter LLMs can effectively learn fine-tuning for accurate sense identification in the presence of ambiguous words. This proposed framework involves sequential tasks: (i) identifying the correct word sense without reasoning, (ii) performing neighbour-word analysis to identify the correct sense, and (iii) applying advanced reasoning that incorporates contextual understanding, justification of the correct sense, and refutation of incorrect senses to improve output quality. 


We propose a novel \textbf{EAD framework} for tasks (ii) and (iii) consisting of three phases: \emph{Exploration (E)}, \emph{Analyzing (A)}, and \emph{Disambiguating (D)}. In the \emph{Exploration} phase, the framework collects sense inventories associated with the ambiguous word, including its interpretations and potential synonyms. The \emph{Analyzing} phase emphasizes reasoning, which involves neighbour word analysis for task (ii) and a deeper evaluation of correct versus incorrect sense interpretations for task (iii). Finally, the \emph{Disambiguating} phase consolidates the outcomes of the reasoning process to determine and retrieve the finalized sense ID. The models selected for the study include Gemma-2-2B~\cite{team2024gemma}, Gemma-3-4B~\cite{team2025gemma}, LLaMA-3.2-1B and LLaMA-3.2-3B~\cite{dubey2024llama}, Qwen-2.5-3B~\cite{yang2025qwen2}, Qwen-3-4B~\cite{yang2025qwen3}, SmoLM-3-3B and DeepSeek-Distill-Qwen1.5B~\cite{liu2024deepseek}. A systematic evaluation of these models was conducted to identify the best candidate to next phase.

\subsubsection{Direct Sense Identification}
\label{direct sense}

To implement the first design objective ``Given a word in a sentence, the LLM must generate the correct definition'', we conducted a supervised evaluation of selected pre-trained language models on their ability to disambiguate the correct meaning of ambiguous words. For this phase, we utilized the FEWS dataset, which contains sentences annotated with target words and their corresponding senses. For instance, one training example includes the sentence: ``He banked the plane sharply to avoid the storm,'' where the word ``bank'' is annotated with the sense ``to tilt or incline an aircraft.'' 
The examples were formatted into instruction-response pairs, where the system prompt specifies the task, the input question provides the sentence and target word, and the output corresponds to the correct sense.

The pre-trained models were first evaluated via inference on these inputs using recommended hyperparameters to assess baseline disambiguation performance. Subsequently, we fine-tuned the baseline models using a supervised instruction fine-tuning approach, following the system prompt-input-output framework described in Subsection~\ref{subsec:study_setup}. Model performance was quantified using BERTScore, precision, recall, and F1 metrics to measure semantic alignment between predicted and reference senses. Initial experiments demonstrated that the Qwen-2.5 3B model exhibited promising learning capabilities after the initial training phase. To further investigate the impact of hyperparameters, additional experiments were conducted by varying the number of epochs and expanding the training dataset with additional SemCor records. These experiments allowed us to identify optimal hyperparameter configurations, which informed the subsequent phases of the study and ensured improved model generalization on the WSD.

\subsubsection{Neighbour Words Analysis\label{cot reasoning}}

The second design objective of ``Given a word in a sentence and possible senses, the model reasons on the neighbouring context closely related to the disambiguation process to identify the correct sense'' was developed in this phase. Inspired by~\citet{guzman2025sandwich}, we employed the fine-tuning design process following an EAD framework, giving full attention to the neighbouring words and following a CoT reasoning process for sense disambiguation. For the dataset creation, we implemented a context-extraction module using a windowed semantic similarity approach. Specifically, each sentence was pre-processed to mark the ambiguous word using <WSD> tags, and spaCy was used to tokenize the preceding and following segments while filtering out stopwords and non-alphabetic tokens. From this, a fixed-size context window of up to 10 tokens on each side was extracted. The target word and its context tokens were then embedded using a sentence-transformer model, and cosine similarity scores were computed between the target embedding and each context token embedding. This allowed the contextual tokens to be ranked by semantic closeness to the target, from which the top-k (default k=5) most semantically relevant words were retained as features. While cosine similarity does not explicitly model syntactic dependencies, it serves as a salience-based heuristic to identify lexically influential neighbouring tokens within a constrained window. Importantly, the full sentence is retained during model fine-tuning, allowing the transformer architecture to capture long-range syntactic relations beyond the similarity-based feature selection. These context-sense pairs were then integrated into the dataset, ensuring that the training data explicitly encoded the most influential neighbouring words for disambiguation. 

To exemplify the process, the sentence ``After the match, the <WSD>bat</WSD> was placed carefully back into the player’s bag'' contains the target ambiguous word `bat' which has 12 distinct noun senses according to the FEWS sense inventory. The preceding context tokens [`After', `match'] and following context tokens [`placed', `carefully', 'back', 'player', 'bag'] were extracted, and similarity scores were computed with `bat'. The tokens were then ranked by their similarity values, and the top-k most semantically related neighbours were selected. In this example, the high similarity of `match', `player', and `bag' to `bat' strongly indicates the sports equipment sense, as opposed to the flying mammal or old woman senses. These ranked neighbours serve as semantically aligned cues from the local context, guiding the disambiguation process toward the correct interpretation.

We applied this process to annotate 101K FEWS and 226K SemCor data records, structuring the output in CoT reasoning, then used this dataset to fine-tune the models.
Our experiments revealed that this approach was more efficient than standard fine-tuning, yielding promising results for disambiguation. However, we observed that LLaMA 1B, LLaMA 3B, and DeepSeek-distill-Qwen 1.5B did not acquire the expected level of reasoning capability. The other models demonstrated strong reasoning performance, showing robust generalization and effective disambiguation even for unseen sentences.

\subsubsection{Advanced Reasoning}
\label{Advanced Reasoning}

This phase was designed to achieve the goal: ``Given a word in a sentence and its possible senses, the model reasons why a particular sense is correct and why the others are not,'' to determine the correct meaning. The main objective of this process is to disambiguate meaning through a deep analysis of each sense in relation to the sentence requiring disambiguation. Due to the lack of an appropriate reasoning-specific dataset, we used a large-parameter open source LLM with a carefully designed prompt. 
We used a dataset of 10K samples which were randomly selected from the FEWS dataset to ensure coverage across all POS tags.

We discovered that three elements are critical to effective reasoning in sense disambiguation:~(i)~contextual analysis,~(ii)~justification of the correct sense, and~(iii)~systematic elimination of incorrect senses. Our design was inspired by~\citet{huang_glossbert_2020}, who reported the importance of using context-gloss pairs to train models not only to identify why a sense ID is valid but also to reject invalid ones. Adopting the setup described in section~\ref{subsec:study_setup}, we fine-tuned the selected model and achieved comparable results despite the small sample size, demonstrating improved reasoning performance for disambiguation. This process aligns with the EAD framework, where exploration gathers candidate senses, analysis provides reasoning for sense selection, and disambiguation yields the final sense ID.

After the initial reasoning process, we observed limitations in verb disambiguation. To address this challenge, we incorporated syntactic evidence into the reasoning process, where a verb’s morphosyntax (tense, aspect, voice), immediate dependents (subjects, objects, complements), key function words (auxiliaries, particles, prepositions), and relevant dependency or constituent patterns were explicitly included in the prompt to constrain its meaning. To evaluate the effect of these modifications, we trained and tested Qwen-3 and Gemma-3 models exclusively on newly constructed training data tailored for this purpose. The evaluation was conducted using the FEWS Few-shot Development set with verb subset, consisting of 2.2K samples, ensuring the assessment was focused on verb disambiguation. Moreover, we employed a hybrid training approach that combined reasoning with neighbour, syntactic, and semantic analysis, and the results were recorded to validate the improvements in learning and disambiguation accuracy.

\subsubsection{Ablation Study}
\label{Ablation Study}

The experimental findings showed that reasoning-based approaches are well-suited for handling unseen data. To evaluate these findings further, we examine the generalizability of such models to different datasets without any additional fine-tuning. 

Thus, we employed the recently released ``Fool Me If You Can'' dataset~\cite{ballout2024fool}, an adversarial benchmark designed to investigate the robustness of language models in WSD. We selected this dataset for the ablation study due to the varied contextual challenges. Specifically, we use the FEWS sense-mapped version retrieved from prior work~\cite{sumanathilaka_glossgpt_2025} \footnote{https://github.com/Sumanathilaka/FOOL-ME-IF-YOU-CAN-dataset-Meets-FEWS-sense-Tags}. The dataset consists of four subsets: a baseline with context-appropriate meanings similar to training data, for standard WSD evaluation; a version adding adjectives that reinforce intended meanings; another replacing these with adjectives linked to the opposite sense; and newly crafted natural sentences where context contradicts the expected meaning, creating harder disambiguation scenarios. Fine-tuning on the ``Fool Me If You Can'' training data was intentionally avoided to ensure that the evaluation measured the robustness and generalization capabilities of the current reasoning models when exposed to unseen data from a different dataset. 

To further probe these capabilities, we also evaluated the models on two additional challenging benchmarks: hardEn~\cite{maru_nibbling_2022}, which comprises 476 test cases that cannot be solved by existing state-of-the-art WSD models, and 42D, a specialized dataset containing 370 records with rare and domain-specific senses as defined by BabelNet~\cite{navigli2010babelnet}.

This design choice allows us to assess whether the models can maintain disambiguation performance under domain shift and adversarial context, rather than benefiting from dataset-specific adaptation (see Section~\ref{res:ablationstudy} for results).




\subsection{Study Setup}
\label{subsec:study_setup}
For model development, supervised fine-tuning (SFT) was employed to adapt a pre-trained LLM to WSD task. The baseline models were fine-tuned using Low-Rank Adaptation (LoRA) to enable efficient training while reducing computational overhead. Training data were pre-processed into a desired chat-style prompt–response format, with each example containing a context sentence and the ambiguous target word to be disambiguated.

Fine-tuning was conducted using Hugging Face transformers and trl libraries, with a custom prompt formatting function to standardize inputs. The datasets were tokenized using the model’s native tokenizer, and training hyperparameters: batch size $=$ 4, gradient accumulation $=$ 8 steps, learning rate $=$ $2 \times 10^{-4}$. We used AdamW 
optimizer and linear learning rate scheduler 
with a random seed of 3407 as recommended by prior work~\cite{picard2023torchmanualseed3407needinfluencerandom}. The fine-tuning used standard supervised causal language modelling, with cross-entropy loss backpropagated through the gold output tokens.

All experiments were performed on an NVIDIA A100-PCIE-40GB GPU, ensuring sufficient memory bandwidth for efficient fine-tuning of the model without quantization in the final setup. The resulting LoRA adapters and tokenizer were saved locally and uploaded to the Hugging Face Hub for reproducibility and deployment, which can be accessed at \url{https://huggingface.co/deshanksuman}.

\section{Results and Discussion}

\begin{table*}[htbp]
\centering
\label{tab:bertscoreintial}
\caption{Bert Score based on distilled base uncased against the base model vs finetuned model. The best scores are in bold. * Deepseek-distill-Qwen 1.5B}
\begin{footnotesize}
\begin{tabular}{p{2.5cm}|ccc|ccc}
\hline
\multirow{2}{*}{\textbf{Models}} & 
\multicolumn{3}{c|}{\textbf{Base Inference }} & 
\multicolumn{3}{c}{\textbf{Instructional Fine-Tuning}} \\ \cline{2-7}
& \textbf{Precision} & \textbf{Recall} & \textbf{F1 Score} 
& \textbf{Precision} & \textbf{Recall} & \textbf{F1 Score} \\ \hline

Gemma 2 2B & 0.7720 & 0.7341 & 0.7515 & 0.7870 & 0.7781 & 0.7814 \\
Gemma 3 4B & \textbf{0.7806} & 0.7302 & 0.7534 & 0.7741 & 0.7214 & 0.7458 \\
Llama 3.2 1B & 0.6439 & 0.7503 & 0.6923 & 0.7919 & 0.7474 & 0.7681 \\
Llama 3.2 3B & 0.7616 & 0.7330 & 0.7430 & \textbf{0.8079} & 0.7567 & 0.7804 \\
SmolLM 3 & 0.7687 & \textbf{0.7656} & \textbf{0.7662} & 0.7884 & 0.7636 & 0.7749 \\
Qwen 2.5 3B & 0.7551 & 0.6773 & 0.7130 & 0.8126 & \textbf{0.8018} & \textbf{0.8061} \\
Qwen 3 4B & 0.7619 & 0.7470 & 0.7531 & \textbf{0.8128} & 0.7970 & 0.8036 \\
Deepseek 1.5B* & 0.7248 & 0.7193 & 0.7212 & 0.7140 & 0.7491 & 0.7294 \\

\hline
\end{tabular}
\end{footnotesize}
\end{table*}

In the initial phase, we evaluated eight different low-parameter LLMs to assess their performance in disambiguating a word within a given sentence. Inspired by~\citet{sumanathilaka2024assessing}, we evaluate using a subset of the FEWS test dataset containing 1,050 cases distributed across nouns, verbs, and adjectives in a 4:3:3 ratio and with 50 adverb examples, to ensure coverage across parts-of-speech.

Performance was measured using BERTScore, computed with the distilled base uncased models. BERTScore was chosen because the evaluation focused on comparing the semantic similarity between the expected meaning and the model’s predicted output, rather than exact token matching. As the initial experiments were not sense-mapped, a meaning-based metric provided a more informative estimate of model performance.

We applied an SFT approach without explicit reasoning steps to examine how well the models could learn the senses of ambiguous words. Fine-tuning revealed clear improvements in sense identification and disambiguation for the Qwen and LLaMA models (see Table~\ref{tab:bertscoreintial}). Gemma-3-4B showed a performance drop in the sense identification setting. However, this was not due to poor disambiguation; rather, the model tended to generate longer, more detailed sense explanations rather than concise labels. Because evaluation relied on BERTScore against short reference descriptions, the resulting outputs were more verbose, leading to lower similarity scores. Notably, this issue did not appear in the sense ID prediction setting with predefined candidates.

Given the better performance of Qwen-2.5-3B, we further examined the effect of the number of training epochs. We found that increasing the number of epochs from one to five did not yield significant improvements in performance for F1 score (1:0.568, 3: 0.552, 5:0.571). McNemar tests~\cite{mcnemar1947note} did not reveal any statistically significant differences. Thus, we limited fine-tuning to two epochs to avoid unnecessary computation overhead.

\subsection{Reasoning Experiments}

In the initial experiments of the second phase, we fine-tuned the eight models to evaluate their ability to learn reasoning using neighbouring word analysis. All models were initially trained for 1 epoch to assess performance and selected models were extended to 2 epochs based on performance. Notably, Llama 1B, 3B, and Deepseek-distill-Qwen 1.5B did not learn reasoning effectively and behaved primarily as text predictors, failing to follow the instructions given during inference.

As presented in Table~\ref{tab:finetuningCOT}, fine-tuning significantly improves performance over the baseline. For instance, Qwen-3-4B and Gemma-3-4B models showed substantial gains after 2 epochs, achieving the highest overall scores of 0.738 and 0.752, respectively. Among the parts of speech, Gemma-3-4B achieved the highest score in Adverbs (0.80) after 1 epoch, while Qwen-3-4B showed the best improvement in Adjectives (0.75) after 2 epochs. These observations indicate that fine-tuning leads to consistent improvements across all categories, confirming that small models with reasoning-based fine-tuning benefit more from extended training compared to baselines.

\begin{table*}[htbp]
\centering
\caption{Fine Tuning with CoT based reasoning with Neighbour words analysis. The results are presented in the F1 score.}
\label{tab:finetuningCOT}
\begin{footnotesize}
\begin{tabular}{l|ccccc}
\hline
\textbf{Models} & \textbf{Noun} & \textbf{Verb} & \textbf{Adjective} & \textbf{Adverb} & \textbf{Overall} \\ \hline

\multicolumn{6}{l}{\textbf{Fine Tuned with CoT based Neighbour words analysis Approach}} \\
\hline
SmolLM 3B (1 epoch) & 0.52 & 0.35 & 0.50 & 0.54 & 0.47 \\
Gemma-2 2B (1 epoch) & 0.70 & 0.61 & 0.68 & 0.66 & 0.67 \\
Qwen 2.5 3B (1 epoch) & 0.71 & 0.64 & 0.71 & 0.74 & 0.69 \\
Gemma 3 4B (1 epoch) & 0.79 & 0.65 & 0.70 & \textbf{0.80} & 0.72 \\
Qwen 3 4B (2 epochs) & 0.79 & 0.67 & \textbf{0.75} & 0.68 & 0.74 \\
Gemma 3 4B (2 epochs) & \textbf{0.81} & \textbf{0.71} & 0.72 & 0.76 & \textbf{0.75} \\
\hline
\multicolumn{6}{l}{\textbf{Current Baseline}} \\
\hline
Qwen 3 4B  & 0.69 & 0.54 & 0.58 & 0.48 & 0.61 \\
Gemma 3 4B & 0.62 & 0.48 & 0.52 & 0.48 & 0.54\\
Gemma 7B & 0.49 & 0.41 & 0.51 & 0.46 & 0.47 \\
Mixtral 7B & 0.43 & 0.32 & 0.46 & 0.42 & 0.41 \\
Yi - 34B & 0.65 & 0.51 & 0.57 & 0.52 & 0.58 \\
GPT 4o-mini & 0.37 & 0.30 & 0.31 & 0.32 & 0.33\\

\hline
\end{tabular}
\end{footnotesize}
\begin{flushleft}
\end{flushleft}
\end{table*}

To evaluate the effectiveness of different reasoning strategies, we benchmark the CoT reasoning-based approach against the few-shot reasoning based approach and the advanced reasoning with correct and incorrect sense analysis. For the few-shot reasoning based approach, we employed the same fine-tuned models; however, during inference, we provided additional few-shot examples to further emphasize understanding of the target senses, in line with evidence that supplementary contextual examples could improve performance~\cite{yang2024multiple}. Each sense was accompanied by two illustrative examples that demonstrated its meaning and interpretation within a specific context. For evaluation, the dynamic few-shot examples were retrieved from a pre-constructed knowledge base.

Table~\ref{tab:finetining_benchmark} indicates that the zero-shot CoT reasoning approach generally outperforms the few-shot reasoning approach in terms of overall accuracy, particularly for the Gemma-3-4B model fine-tuned over two epochs (0.75). Interestingly, this zero-shot CoT configuration also achieves the highest noun (0.81) and verb (0.71) scores across all tested settings. While the few-shot approach provides additional contextual cues through example cases, it does not consistently translate into higher performance, suggesting that the models are already able to leverage their fine-tuned reasoning capabilities without supplementary examples. The advanced reasoning with correct and incorrect sense analysis approach performs competitively, with Qwen-3-4B reaching strong adverb performance (0.80) and stable results across categories, but still not surpassing the zero-shot CoT reasoning in overall accuracy. Importantly, the advanced reasoning approach used only 10\% of the training data compared to the CoT reasoning approach, yet produced comparable results. This trend underscores the robustness of the CoT reasoning paradigm, even in the absence of explicit few-shot demonstrations, and highlights its ability to generalize reasoning patterns effectively across different POS categories.

\begin{table*}[htbp]
\centering
\caption{F1 Score for benchmarking Different Reasoning Strategies. 
}
\label{tab:finetining_benchmark}
\begin{footnotesize}
\begin{tabular}{l|ccccc}
\hline
\textbf{Models} & \textbf{Noun} & \textbf{Verb} & \textbf{Adjective} & \textbf{Adverb} & \textbf{Overall} \\ \hline

\multicolumn{6}{l}{\textbf{Fine Tuned with COT based Neighbour words analysis Approach (Zero shot)}} \\
\hline
Qwen 3 4B (2 epochs) & 0.79 & 0.67 & 0\textbf{.75} & 0.68 & 0.74$^{**}$ \\
Gemma 3 4B (2 epochs) & \textbf{0.81 }& \textbf{0.71} & 0.72 & \textbf{0.76} & \textbf{0.75}$^{**}$\\

\hline
\multicolumn{6}{l}{\textbf{Fine Tuned with COT based Neighbour words analysis Approach (Few shot)}} \\
\hline
Qwen 3 4B (2 epochs) & 0.74 & 0.62 & 0.68 & 0.64 & 0.68 \\
Gemma 3 4B (2 epochs) & \textbf{0.78} & \textbf{0.68} &\textbf{ 0.70} & \textbf{0.78} & \textbf{0.72} \\

\hline
\multicolumn{6}{l}{\textbf{Advanced Reasoning with Correct and Incorrect Sense Analysis}} \\
\hline
Gemma 3 4B (2 epochs) & 0.76 & 0.60 & 0.66 & 0.64 & 0.68 \\
Qwen 2.5 3B (2 epochs) & 0.74 & 0.61 & 0.68 & 0.66 & 0.68 \\
Qwen 3 4B (2 epochs) & \textbf{0.76} & \textbf{0.67 }& \textbf{0.70 }& \textbf{0.80} &\textbf{ 0.72} \\

\hline

    \multicolumn{6}{r}{***$p<0.001$, **$p<0.01$, *$p<0.05$}

\end{tabular}
\end{footnotesize}
\end{table*}

To gain a broader understanding of the performance of the proposed reasoning-based approach compared to low-parameter models, we evaluated on the FEWS test set, which contains both few-shot and zero-shot datasets. Notably, the few-shot setting included examples whose senses do appear in the training set, but only in limited numbers, whereas the zero-shot setting contained examples of senses not encountered during training. We benchmarked our results against current state-of-the-art models to assess the effectiveness of our approach. Thus, we compared against MFS, Lesk, SemEq~\cite{yao_connect--dots_2021}, ESR~\cite{song_improved_2021}, RTWE~\cite{zhang_word_2023}, and GlossGPT~\cite{sumanathilaka_glossgpt_2025} (see Table~\ref{tab:fewstesteval}).

The proposed models (Qwen \& Gemma) show competitive performance despite having significantly fewer parameters than most competitor systems. With the Few-shot set, both models outperform traditional baselines (i.e. MFS \& Lesk) by a considerable margin and achieve close to state-of-the-art results. Notably, Qwen achieves 76.52 and Gemma 75.68, surpassing RTWE\textsubscript{L} and approaching GlossGPT’s performance. In the more challenging Zero-shot set, the models maintain strong accuracy (72.66 and 71.86), outperforming most baselines, showing robust generalization. The results show the effectiveness of our reasoning-based fine-tuning approach, even using compact models.

\begin{table*}[!t]
\centering
\caption{F1 score for CoT based Neighbour words analysis against current techniques. FEWS test data used for evaluation. \textsubscript{b}: Base, \textsubscript{L}: Large, Our approach in italic (4B models)}
\label{tab:fewstesteval}
\begin{footnotesize}
\begin{tabular}{l|c|c|c|c|c|c|c|c|c|c}
\hline
Dataset & MFS & Lesk & SemEq\textsubscript{b} & SemEq\textsubscript{L} & ESR\textsubscript{b} & ESR\textsubscript{L} & RTWE\textsubscript{L} & \textit{Qwen} & \textit{Gemma} & GlossGPT \\ \hline

Fewshot& 51.5 & 40.9 & 80.1 & 82.3 & 77.8 & 83.4 & 78.4 & \textit{76.52} & \textit{75.68} & \textbf{90.7} \\ 
Zeroshot& - & 39.0 & 70.2 & 72.2 & 71.6 & 75.8 & 69.9 & \textit{72.66} & \textit{71.86 }& \textbf{79.5} \\ 

\hline
\end{tabular}
\end{footnotesize}

\end{table*}

\begin{table*}[!t]
\centering
\caption{
F1 score of the VERB task compared using different reasoning methods. Hybrid:Neighbour Word Analysis with Syntactic \& Semantic Analysis, A: Analysis}
\label{tab:verb_reasoning_comparison}
\begin{footnotesize}
\begin{tabular}{ccccc}
\hline
\textbf{Model} & \textbf{Neighbour Word A.} & \textbf{Semantic A. } & \textbf{Syntactic \& Semantic A.} & \textbf{Hybrid} \\
\hline
Qwen3-4B  & \textbf{0.679}$^{*}$ & 0.662 & 0.677 & 0.602 \\
Gemma3-4B & \textbf{0.658}$^{**}$ & 0.601 & 0.628 & 0.544\\
\hline
 \multicolumn{5}{r}{***$p<0.001$, **$p<0.01$, *$p<0.05$}
\end{tabular}
\end{footnotesize}
\end{table*}

From Table~\ref{tab:finetining_benchmark}, we observed that verb disambiguation remains challenging for all models. We explored different fine-tuning strategies to test model reasoning improvements (see setup in section~\ref{Advanced Reasoning}). 
However, verb-specific models still struggle to disambiguate verbs using syntactic clues, even when fine-tuned or combined in a hybrid model that incorporates both reasoning and neighbour analysis with syntactic and semantic features. 
Table~\ref{tab:verb_reasoning_comparison} shows that adding extra reasoning signals still fails due to being interpreted as noise.



\subsection{Ablation Study}

\label{res:ablationstudy}

\begin{table*}[t!]
\centering
\caption{Performance comparison (F1 score) against existing WSD models. The compared models are ARES~\cite{scarlini-et-al-2020-ares},BEM~\cite{blevins_moving_2020}, ESC~\cite{barba_esc_2021}, EWS~\cite{bevilacqua2020breaking}, GEN~\cite{bevilacqua2020generationary}, SYN~\cite{scozzafava-etal-2020-personalized}),CSC~\cite{barba_consec_2021} and SandWiCH~\cite{guzman2025sandwich}. Our best-performing approach, Neighbour words analysis with Finetuned Qwen and Gemma, is reported in Italic.}

\label{tab:baseline_comparison}
\begin{tabular}{lccccccccccc}
\hline
\textbf{Dataset} & \textbf{ARES} & \textbf{BEM} & \textbf{ESC} & \textbf{EWS} & \textbf{GEN} & \textbf{SYN} & \textbf{CSC} & \textbf{SandWiCH} & \textbf{\textit{Gemma}} & \textbf{\textit{Qwen}}\\
\hline
42D     & 41.8 & 53.2 & 58.9 & 43.9 & 50.2 & 32.8 & 56.6 & 77.1 &64.43& \textbf{78.48}\\
hardEN  & 0.0  & 0.0  & 0.0  & 0.0  & 0.0   & 0.0  & 7.35 & 53.4 & 40.71 & \textbf{54.19}\\
\hline
\end{tabular}
\end{table*}

\begin{table*}[t!]
\centering
\caption{F1 score for ``Fool me if you can'' dataset for binary classification of sense ID. *Our approach is not fine-tuned with training data from ``Fool me if you can'' dataset.}
\label{tab:ablationstudy}
\begin{footnotesize}
\begin{tabular}{l|ccccc}
\hline
\textbf{Models} & \textbf{\# Parameters} & \textbf{Set 1} & \textbf{Set 2} & \textbf{Set 3} & \textbf{Set 4} \\ \hline

Roberta-base & 125M & 0.945 & 0.969 & 0.888 & 0.715 \\
Bert-large & 340M & 0.970 & 0.978 & 0.874 & 0.689 \\
T5-large & 770M & 0.984 & 0.987 & 0.896 & 0.691 \\
FLAN-T5-large & 780M & 0.948 & 0.953 & 0.852 & 0.663 \\
T5-xl & 3B & \textbf{0.991} & 0.992 & \textbf{0.907} & 0.710 \\
FLAN-T5-xl & 3B & 0.955 & 0.958 & 0.881 & 0.718 \\
Mixtral 7bx8 & 7B & 0.987 & \textbf{0.993} & 0.820 & 0.714 \\
Llama3 8B & 8B & 0.986 & 0.990 & 0.790 & 0.687 \\


\textit{Gemma 3 (Reasoning)*} & 4B & 0.966 & 0.969 & 0.812 & 0.811 \\
\textit{Qwen 3 (Reasoning)*} & 4B & 0.970 & 0.972 & 0.847 & \textbf{ 0.852} \\

\hline
\end{tabular}
\end{footnotesize}
\end{table*}

The ablation study aimed to evaluate the robustness and generalizability of the trained module by applying CoT based reasoning combined with Neighbour Words Analysis to new, unseen data of two diverse use cases. 

Table \ref{tab:baseline_comparison} presents a detailed comparison of our proposed approach against several state-of-the-art WSD systems on the challenging hard WSD benchmarks introduced by \citet{maru_nibbling_2022}. Our method, leveraging neighbour word analysis combined with Finetuned Qwen and Gemma, demonstrates superior performance, with Qwen-4B achieving the highest F1 scores across both 42D and hardEN datasets. Notably, Qwen-4B attains an F1 score of 78.48 on the hardN benchmark, substantially outperforming existing models, while Gemma also shows strong competitive results.

Table \ref{tab:ablationstudy} shows that the model adapts effectively to unseen data with the Fool dataset, maintaining comparable performance across Sets 1-3. This underscores the extent to which the model can handle novel scenarios and maintain robust reasoning capabilities. Importantly, in the more challenging Set 4, containing realistic sentences where contextual cues deliberately contradict the expected meaning of a homonym, the model reached state-of-the-art performance, even for the small LLM class. This demonstrates the model’s capacity to address difficult disambiguation tasks that require nuanced contextual understanding. When compared to the results reported in the original paper \cite{ballout2024fool}, our approach not only outperformed GPT-3.5-Turbo but also revealed that the key driver of superior performance is not merely model size, but the inclusion of a well-structured reasoning process. This reinforces the importance of targeted reasoning strategies for achieving strong generalization and adaptability in language models. This is an important finding because it challenges the common assumption of larger sized models for improved performance~\cite{yae2025leveraging}. Instead, it reveals that carefully designed reasoning strategies like \textit{CoT combined with Neighbour Words Analysis} can yield state-of-the-art results even with smaller models.

\section{Conclusion and Future Work}

Our study shows that low-parameter LLMs can achieve competitive and state-of-the-art performance in WSD when equipped with reasoning-based fine-tuning strategies. Through evaluations with eight \texttt{<}4B parameter models, we showed that CoT reasoning combined with neighbour-word analysis enables strong sense prediction even in zero-shot and domain-shift settings. Gemma-3-4B and Qwen-3-4B models consistently outperformed medium-parameter baselines and were comparable to larger models (e.g., GPT-3.5-Turbo), while maintaining robustness across adversarial datasets and hard WSD datasets.

Importantly, the proposed approaches were effective in modest computational settings, with the advanced reasoning strategy achieving competitive accuracy using only 10\% of the training data compared to the CoT method. These findings support the fact that reasoning quality is a critical determinant of LLM-based WSD performance. Future work should extend this approach to multilingual settings, including low-resource languages, to assess its adaptability across linguistic contexts. 

\section*{Limitations}

The scope of this study is limited to eight models from different vendors, all of which have fewer than 4B parameters. While further exploration of mid-sized models may yield improved performance, the primary objective, demonstrating performance gains through enhanced reasoning has been successfully achieved and remains adaptable. The current study focuses exclusively on English WSD; however, the approach can be extended to multilingual or cross-lingual settings in future work. Due to computational constraints, training iterations were restricted to two epochs, and only a sample of the generated data was validated owing to limited resource availability. Nevertheless, the evaluated samples produced promising results consistent with the expectations for this study.

\section*{Acknowledgements}

We acknowledge the support of the Super computing Wales project, which is part-funded by the European Regional Development Fund (ERDF) via Welsh Government. Hough's work is supported by the EPSRC grant EP/X009343/1 `FLUIDITY'. 

\section*{Ethics consideration}
This paper has been conducted in compliance with Swansea University's ethical standards. 

\section*{Data \& Code availability}

The fine-tuned LoRA adapters used in this study are publicly available on 
Hugging Face\footnote{\href{https://huggingface.co/deshanksuman}{https://huggingface.co/deshanksuman}}, 
enabling reproducibility and further research. 
In addition, the full training, inference, and evaluation pipeline has been 
released as open-source code on GitHub\footnote{\href{https://github.com/Sumanathilaka/An-EAD-Reasoning-Framework-for-WSD-with-Low-Parameter-LLMs}{https://github.com/Sumanathilaka/An-EAD-Reasoning-Framework-for-WSD-with-Low-Parameter-LLMs}}.

\nocite{*}
\section*{Bibliographical References}\label{sec:reference}

\bibliographystyle{lrec2026-natbib}
\bibliography{lrec2026-example}

\label{lr:ref}
\bibliographystylelanguageresource{lrec2026-natbib}
\bibliographylanguageresource{languageresource}

\newpage

\section{Appendix}







\begin{table}[!h]
\begin{footnotesize}
\centering
\caption{Prompt used to elicit rationales for sense selection during fine-tuning.}
\label{tab:cot-prompt}
\begin{tabular}{p{7.5cm}}
\hline
You are a helpful assistant tasked with simulating the human thinking process for Word Sense Disambiguation (WSD).
Given a sentence containing an ambiguous word, along with the expected sense ID, your goal is to generate a detailed and logical reasoning process as a human would.

Your response should include:
\begin{itemize}[itemsep=0pt]
    \item Contextual Analysis – Explain how the surrounding context in the sentence helps determine the correct meaning of the ambiguous word.
\item Justification of the Correct Sense ID – Provide a clear explanation of why the expected sense ID is appropriate in this context.
\item Elimination of Incorrect Senses – Briefly describe why the other possible sense IDs do not fit the given context.
\end{itemize}
Be thorough, logical, and emulate how a human would naturally think through the disambiguation. Do not include any additional information or instructions outside of the reasoning process.
\\
\hline

\end{tabular}
\end{footnotesize}
\end{table}

\begin{table}[H]
\begin{footnotesize}
\centering
\caption{Prompt used to elicit rationales for Verb sense selection during extended study.}
\label{tab:cot-prompt-verb}
\begin{tabular}{p{7.5cm}}
\hline
You are a helpful assistant tasked with simulating the human thinking process for Word Sense Disambiguation (WSD).

Given a sentence containing an ambiguous word, along with the expected sense ID, your goal is to generate a detailed and logical reasoning process as a human would.

The given ambiguous word is a VERB.

Your response should include:

\begin{enumerate}[itemsep=0pt]

 \item  Syntactic Evidence — Summarize the verb’s morphosyntax (tense/aspect/voice), immediate dependents (subject, objects, complements), key function words (auxiliaries, particles, prepositions), and any relevant dependency/constituent patterns that constrain its meaning.

 \item  Semantic Evidence — Describe selectional preferences and semantic roles implied by the verb, plausible paraphrases, collocations, and context/topic cues (entities, events) that support a specific sense.

 \item  Decision — State the chosen sense ID and give a clear justification that ties the syntactic and semantic cues to that sense.

 \item  Elimination of Alternatives — Briefly state why other possible sense IDs do not fit given the observed syntax and semantics.

\end{enumerate}
Be thorough, logical, and emulate how a human would naturally think through the disambiguation. Do not include any additional information or instructions outside of the reasoning process.
\\
\hline

\end{tabular}
\end{footnotesize}
\end{table}

\begin{table*}[t]
\centering
\caption{Prompt used for LLM-as-a-Judge evaluation of WSD reasoning quality}
\label{tab:cot-prompt-LLM-as-judge}
\begin{tabular}{p{\linewidth}}
\hline
You are an expert evaluator assessing the quality of Word Sense Disambiguation (WSD) reasoning generated by an AI system.

\textbf{Task Context:}
\begin{itemize}
    \item Original sentence, ambiguous word, and sense definitions: \{input\_text\}
    \item Correct sense ID: \{senseid\}
    \item Generated Reasoning to evaluate: \{reasoning\}
\end{itemize}

\textbf{Evaluation Instructions:}

Evaluate the generated reasoning on the following four dimensions using a 1-5 scale:

\begin{enumerate}
    \item \textbf{Contextual Analysis Quality}: Does the reasoning identify and explain relevant contextual clues effectively?
    \begin{itemize}
        \item 5: Comprehensively identifies all relevant contextual clues and explains their disambiguation role
        \item 4: Identifies most key contextual elements with clear explanations
        \item 3: Identifies some context but misses important clues or lacks depth
        \item 2: Minimal contextual analysis with superficial observations
        \item 1: Incorrect or irrelevant contextual analysis
    \end{itemize}

    \item \textbf{Sense ID Justification Accuracy}: Is the justification for the correct sense logically sound and semantically accurate?
    \begin{itemize}
        \item 5: Provides precise, logically sound justification with clear semantic connections
        \item 4: Good justification with minor gaps in reasoning
        \item 3: Adequate justification but lacks depth or contains minor logical errors
        \item 2: Weak justification with significant logical gaps
        \item 1: Incorrect or nonsensical justification
    \end{itemize}

    \item \textbf{Elimination Reasoning Completeness}: Does it systematically eliminate alternative senses with clear reasoning?
    \begin{itemize}
        \item 5: Systematically eliminates all alternative senses with clear reasoning
        \item 4: Eliminates most alternatives with good reasoning
        \item 3: Partial elimination with some reasoning gaps
        \item 2: Minimal elimination or weak reasoning
        \item 1: No elimination or incorrect reasoning
    \end{itemize}

    \item \textbf{Overall Coherence and Human-likeness}: Does the reasoning flow naturally like human thinking?
    \begin{itemize}
        \item 5: Reads like natural human reasoning, logically structured and comprehensive
        \item 4: Good flow with minor artificiality
        \item 3: Adequate but somewhat mechanical or repetitive
        \item 2: Disjointed or overly formulaic
        \item 1: Incoherent or completely artificial
    \end{itemize}
\end{enumerate}

\textbf{Output Format (JSON):}

\begin{verbatim}
{
  "contextual_analysis_score": 1-5,
  "justification_accuracy_score": 1-5,
  "elimination_completeness_score": 1-5,
  "coherence_score": 1-5
}
\end{verbatim}

\end{tabular}
\end{table*}

\end{document}